\documentclass{article}

\usepackage{arxiv}

\newcommand{\code}[1]{\texttt{#1}}

\usepackage{amsmath}
\usepackage{amssymb}   

\usepackage{array}
\usepackage{booktabs}  

\usepackage[table]{xcolor}
\definecolor{codebg}     {HTML}{F8F8F8}
\definecolor{codeframe}  {HTML}{DDDDDD}
\definecolor{codegreen}  {HTML}{2E7D32}
\definecolor{codeblue}   {HTML}{1565C0}
\definecolor{codepurple} {HTML}{6A1B9A}
\definecolor{codegray}   {HTML}{9E9E9E}

\usepackage{enumitem}
\usepackage{siunitx}

\usepackage{graphicx}
\usepackage{tikz}
\usetikzlibrary{
    positioning,
    shapes.geometric,
    shapes.misc,
    arrows.meta,
    calc,
    mindmap,
    shadows,
    fit,
    backgrounds,
    patterns
}

\usepackage{pgfplots}
\pgfplotsset{compat=1.18}

\usepackage{listings}
\usepackage{eurosym}               

\usepackage[numbers,sort&compress]{natbib}   

\usepackage{hyperref}
\hypersetup{
  colorlinks = true,
  linkcolor  = codeblue,
  citecolor  = codeblue,
  urlcolor   = codeblue,
}
\usepackage{url}

\usepackage[acronym, nogroupskip, nonumberlist, nopostdot]{glossaries}
\glsdisablehyper

\providecommand{\Description}[1]{}

\newenvironment{acks}{%
  \section*{Acknowledgements}%
}{\par}

\title{Detecting Ladder Logic Bombs in IEC~61131-3 PLC Programs using
ESBMC-PLC+: A Formal Verification Approach with Trigger Synthesis}

\renewcommand{\shorttitle}{Detecting Ladder Logic Bombs in IEC~61131-3 PLC Programs}

\renewcommand{\headeright}{Preprint}
\renewcommand{\undertitle}{Preprint}

\author{%
  Pierre Dantas\thanks{The authors contributed equally to this research.
  Corresponding author: \texttt{pierre.dantas@manchester.ac.uk}.} \\
  Department of Computer Science\\
  The University of Manchester\\
  Manchester, UK\\
  \texttt{pierre.dantas@manchester.ac.uk}\\
  \texttt{ORCID: 0000-0001-6390-9340}
  \And
  Lucas Cordeiro\footnotemark[1] \\
  Department of Computer Science\\
  The University of Manchester\\
  Manchester, UK\\
  \texttt{lucas.cordeiro@manchester.ac.uk}\\
  \texttt{ORCID: 0000-0002-6235-4272}
  \And
  Waldir Junior\footnotemark[1] \\
  Electrical Engineering\\
  Federal University of Amazonas (UFAM)\\
  Manaus, AM, Brazil\\
  \texttt{waldirjr@ufam.edu.br}\\
  \texttt{ORCID: 0000-0003-3095-0042}
}

\date{July 2026}

\begin{document}

\newacronym{aadl}{AADL}{Architecture Analysis and Design Language}
\newacronym{ansi-c}{ANSI-C}{American National Standards Institute C}
\newacronym{api}{API}{Application Programming Interface}
\newacronym{ast}{AST}{Abstract Syntax Tree}
\newacronym{bdd}{BDD}{Binary Decision Diagrams}
\newacronym{bmc}{BMC}{Bounded Model Checking}
\newacronym{cbmc}{CBMC}{Bounded Model Checking for ANSI-C Programs}
\newacronym{cegar}{CEGAR}{Counterexample-Guided Abstraction Refinement}
\newacronym{cern}{CERN}{Conseil Européen pour la Recherche Nucléaire}
\newacronym{cfg}{CFG}{Control Flow Graph}
\newacronym{chc}{CHC}{Constrained Horn Clause}
\newacronym{cli}{CLI}{Command-Line Interface}
\newacronym{cpu}{CPU}{Central Processing Unit}
\newacronym{ctl}{CTL}{Computation Tree Logic}
\newacronym{cuda}{CUDA}{Compute Unified Device Architecture}
\newacronym{cve}{CVE}{Common Vulnerability and Exposure}
\newacronym{dfs}{DFS}{Depth-First Search}
\newacronym{dsl}{DSL}{Domain-Specific Language}
\newacronym{epsrc}{EPSRC}{Engineering and Physical Sciences Research Council}
\newacronym{esbmc}{ESBMC}{Efficient SMT-based Context-Bounded Model Checker}
\newacronym{fbd}{FBD}{Functional Block Diagram}
\newacronym{fb}{FB}{Functional Block}
\newacronym{fpga}{FPGA}{Field-Programmable Gate Array}
\newacronym{ic3}{IC3}{Incremental Construction of Inductive Clauses for Indubitable Correctness}
\newacronym{ide}{IDE}{Integrated Development Environment}
\newacronym{ml}{ML}{Machine Learning}
\newacronym{iec}{IEC}{International Electrotechnical Commission}
\newacronym{ieee}{IEEE}{Institute of Electrical and Electronics Engineers}
\newacronym{ics}{ICS}{Industrial Control Systems}
\newacronym{il}{IL}{Instruction List}
\newacronym{iot}{IoT}{Internet of Things}
\newacronym{ir}{IR}{Intermediate Representation}
\newacronym{iso}{ISO}{International Organization for Standardization}
\newacronym{ld}{LD}{Ladder Diagram}
\newacronym{llb}{LLB}{Ladder Logic Bombs}
\newacronym{llm}{LLM}{Large Language Model}
\newacronym{ltl}{LTL}{Linear Temporal Logic}
\newacronym{nasa}{NASA}{National Aeronautics and Space Administration}
\newacronym{pdr}{PDR}{Property Directed Reachability}
\newacronym{plc}{PLC}{Programmable Logic Controller}
\newacronym{por}{POR}{Partial Order Reduction}
\newacronym{rtos}{RTOS}{Real-Time Operating System}
\newacronym{sat}{SAT}{Boolean Satisfiability}
\newacronym{scl}{SCL}{Structured Control Language}
\newacronym{sfc}{SFC}{Sequential Function Chart}
\newacronym{slr}{SLR}{Systematic Literature Review}
\newacronym{smt}{SMT}{Satisfiability Modulo Theories}
\newacronym{smtlib2}{SMT-LIB}{Satisfiability Modulo Theories Library}
\newacronym{ssa}{SSA}{Static Single Assignment}
\newacronym{st}{ST}{Structured Text}
\newacronym{stl}{STL}{Statement List}
\newacronym{svcomp}{SV-COMP}{Competition on Software Verification}
\newacronym{tacas}{TACAS}{Tools and Algorithms for the Construction and Analysis of Systems}
\newacronym{ufam}{UFAM}{Federal University of Amazonas}
\newacronym{ukri}{UKRI}{UK Research and Innovation}
\newacronym{pou}{POU}{Program Organization Unit}
\newacronym{tsv}{TSV}{Trusted Safety Verifier}

\maketitle

\begin{abstract}
A \gls{llb} is malicious control logic inserted into a \gls{plc} program that lies dormant until a \emph{trigger} fires a \emph{payload} which manipulates actuators, forges sensor readings, or denies operator control. We make a structural observation about real \gls{llb} datasets: the malicious logic hides \emph{inside function-block bodies}, which existing \gls{ld} verifiers drop from their \gls{ir}, rendering the malicious and benign programs indistinguishable to the prover. We present \textbf{ESBMC-LLB}, a method that \emph{uses ESBMC-PLC+} (an existing IEC~61131-3 verifier) as its verification engine, adding a modeling layer that exposes the function-block-resident logic to the prover and recasts \gls{llb} detection as formal verification: a \emph{scan-watchdog} exposes denial-of-control (non-termination) payloads and \emph{output wiring} exposes actuator-forgery payloads as checkable safety violations. \textit{k}-induction yields an \emph{unbounded proof of bomb-absence} across all scan cycles, and the bounded model checker returns a counterexample that \emph{is the detonation trigger} -- two guarantees that signature-, anomaly-, and \gls{cfg}-triage detectors do not provide. On the public~\citet{Iacobelli2024} dataset, ESBMC-LLB detects all 30~bombs and recovers every trigger; because it reasons over program semantics rather than syntactic patterns, it also detects adaptive triggers (computed, opaque-arithmetic, multi-scan) that evade a \gls{cfg}-triage heuristic. We further report the first evaluation of a semantic model checker on the state-of-the-art's real analog benchmark (PLC-Defuser's SWaT corpus): our analog modeling extension makes the full corpus parseable for the first time and, on the archived v1.0.0 release, detects $149/150$ bombs ($99\%$) with zero false positives, recovering each trigger; on a later corpus version that adds \emph{nonlinear} non-termination bombs, detection drops to $49\%$ as the \gls{smt} backend times out. We conclude that semantic model checking and \gls{cfg}-triage are \emph{complementary} -- the former adds unbounded bomb-absence proofs and adaptive-trigger robustness and handles Boolean/integer and linear analog logic, the latter leads on nonlinear analog non-termination -- and we delineate exactly where each wins.
\end{abstract}


\glsresetall

\section{Introduction}
\label{sec:intro}

\glspl{plc} run the control logic of safety-critical \gls{ics}: water treatment, power, chemical, and manufacturing plants. An attacker who gains write access to a \gls{plc} -- through the engineering workstation, a compromised project file, or the supply chain -- can insert a \emph{\gls{llb}}: a fragment of control logic that remains dormant until a \emph{trigger} fires, then executes a \emph{payload} that sabotages the physical process~\cite{Govil2017}. Unlike IT malware, an \gls{llb} is an ordinary ladder or \gls{st} code; it passes commissioning tests and short simulations because its trigger is engineered to remain inactive during normal operation. ~\citet{Govil2017} introduced the concept and a taxonomy of \gls{llb} triggers and payloads; subsequent work proposed detection architectures built on formal verification~\cite{Iacobelli2024} and industrial deployment~\cite{Bruttomesso2024}.

This paper makes that connection concrete and complete. We build on \textbf{ESBMC-PLC+}~\cite{ESBMCpr5400} -- the unified IEC~61131-3 verifier that accepts textual \gls{ld}, graphical \gls{ld}, and \gls{st}/\gls{scl} through a single \gls{esbmc} backend with \textit{k}-induction -- and observe that the machinery built for \emph{safety} verification is, with no backend change, an \emph{\gls{llb} detector}. We call the resulting method \textbf{ESBMC-LLB}:

\begin{itemize}[leftmargin=1.5em]
    \item An \gls{llb} payload necessarily drives the program into a state that violates a safety or integrity property. Detecting the bomb, therefore, reduces to checking that property.
    \item \textit{k}-induction discharges the property as an \emph{unbounded} proof: a SAFE verdict certifies that \emph{no} reachable scan-cycle state detonates a bomb of the modeled class -- a guarantee signature scanners cannot give.
    \item When a bomb is present, \gls{bmc} returns a counterexample trace. That trace \emph{is the trigger}: the exact input/cycle sequence that arms and fires the payload. Detection thus yields automatic \textbf{trigger synthesis}.
\end{itemize}

A stealthy and widely cited \gls{llb} class is the counter/timer time-bomb (``detonate after $N$ cycles'')~\cite{Govil2017}: an integer-state regime where \gls{bdd}-based unbounded provers are known to suffer state-space explosion, whereas the \gls{smt} bit-vector engine ESBMC-LLB builds on reasons about wide integer state symbolically~\cite{ESBMCpr5400}. We use this property to recover counter/timer detonation triggers; we do \emph{not} re-measure the backend here, and we note up front (\S\ref{sec:rq-swat}) that this advantage does \emph{not} extend to \emph{nonlinear} integer state (e.g.\ a non-terminating loop updating \code{i := i*i}), which remains hard for the \gls{smt} backend and is where a \gls{cfg}-triage detector retains the edge.

\subsection{Contributions}

\begin{enumerate}
    \item \textbf{A modeling layer that exposes \gls{fb}-resident bombs to ESBMC-PLC+} (\S\ref{sec:impl}): We show that \gls{llb}s in the reference dataset hide \emph{inside} \gls{ld} function-block bodies, which existing \gls{ld} verifiers drop from the \gls{ir}; our modeling layer translates each function-block body and feeds it to ESBMC-PLC+ (whose verification engine is used unchanged), making the bomb logic reachable to the prover.
    \item \textbf{Two detection mechanisms with trigger synthesis}: a \emph{scan-watchdog} that turns trigger-gated non-termination (denial-of-control) bombs into reachable safety violations, and \emph{output wiring} that turns value/actuator-forgery bombs into safety-property violations; in both, the \gls{bmc} counterexample recovers the detonation trigger, and \textit{k}-induction certifies bomb-absence unboundedly.
    \item \textbf{Adaptive-adversary robustness} (\S\ref{sec:rq-adaptive}): because it evaluates program semantics rather than syntactic patterns, ESBMC-LLB detects triggers (computed, opaque-arithmetic, multi-scan) engineered to evade \gls{cfg}-based trigger triage -- a class such detectors structurally miss.
    \item \textbf{An honest evaluation, including a real head-to-head and a negative result} (\S\ref{sec:experiments}): $30/30$ detection on the public~\citet{Iacobelli2024} dataset, \SI{100}{\percent} recall with zero false positives on a 310-program Boolean/integer corpus (median \SI{70}{\milli\second}), and no regression ($13/13$). We further present the first run of a semantic model checker on PLC-Defuser's real SWaT benchmark: our analog modeling extension makes the corpus parseable and, on the archived v1.0.0 release, detects $149/150$ bombs (\SI{99}{\percent}) with zero false positives, while on a later corpus version that adds \emph{nonlinear} non-termination bombs detection drops to \SI{49}{\percent} as the \gls{smt} backend times out.
    \item \textbf{A characterization of the semantic-vs-\gls{cfg} trade-off} (\S\ref{sec:discussion}): semantic model checking leads on adaptive triggers and unbounded guarantees over Boolean/integer logic; \gls{cfg}-triage leads on nonlinear analog non-termination. The two are \emph{complementary}, not competing.
\end{enumerate}

\section{Background and Related Work}
\label{sec:background}

\subsection{The \gls{plc} Scan Cycle and ESBMC-PLC+}
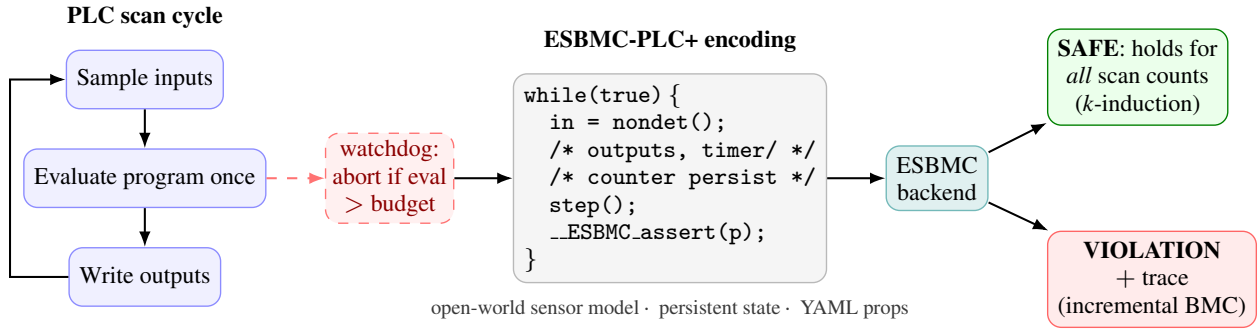
\begin{figure}[htbp]
\centering
\resizebox{\columnwidth}{!}{%
\begin{tikzpicture}[
  font=\scriptsize,
  >={Latex[length=1.6mm]},
  node distance=4mm and 12mm,
  box/.style={draw, rounded corners, align=center, inner sep=3pt, minimum height=6mm},
  phys/.style={box, fill=blue!6,  draw=blue!45},
  code/.style={box, fill=gray!8,  draw=gray!55, align=left, font=\ttfamily\scriptsize},
  eng/.style={box,  fill=teal!12, draw=teal!55},
  ok/.style={box,   fill=green!8, draw=green!50!black},
  bad/.style={box,  fill=red!8,   draw=red!60},
  note/.style={font=\tiny, text=gray!45!black, align=left},
  a/.style={->, semithick}
]
\node[phys] (in)  {Sample inputs};
\node[phys, below=of in]  (ev)  {Evaluate program once};
\node[phys, below=of ev]  (out) {Write outputs};
\draw[a] (in)  -- (ev);
\draw[a] (ev)  -- (out);
\draw[a] (out.west) -- ++(-6mm,0) |- (in.west);   
\node[draw=red!55, dashed, rounded corners, fill=red!6, inner sep=2pt,
      align=center, text=red!55!black, right=6mm of ev] (wd)
      {watchdog:\\ abort if eval\\ $>$ budget};
\draw[a, red!55, dashed] (ev) -- (wd);
\node[font=\bfseries\scriptsize, above=1mm of in] {PLC scan cycle};

\node[code, right=6mm of wd] (enc)
{while(true)\,\{\\
\ \ in = nondet();\\
\ \ /* outputs, timer/ */\\
\ \ /* counter persist  */\\
\ \ step();\\
\ \ \_\_ESBMC\_assert(p);\\
\}};
\node[font=\bfseries\scriptsize, above=1mm of enc] {ESBMC-PLC+ encoding};
\node[note, below=0.5mm of enc] {open-world sensor model\,$\cdot$\, persistent state\,$\cdot$\, YAML props};
\draw[a] (wd.east) -- ++(3mm,0) |- (enc.west);

\node[eng, right=6mm of enc] (be) {\acrshort{esbmc}\\ backend};
\draw[a] (enc) -- (be);
\node[ok,  above right=2mm and 6mm of be] (safe) {\textbf{SAFE}: holds for\\ \emph{all} scan counts\\ (\textit{k}-induction)};
\node[bad, below right=2mm and 6mm of be] (viol) {\textbf{VIOLATION}\\ $+$ trace\\ (incremental \acrshort{bmc})};
\draw[a] (be) -- (safe);
\draw[a] (be) -- (viol);
\end{tikzpicture}}
\caption{The \gls{plc} scan cycle and its ESBMC-PLC+ encoding. A \gls{plc} repeatedly samples inputs, evaluates the program once, and writes outputs, with a watchdog aborting an over-budget evaluation. ESBMC-PLC+ encodes the cycle as a \code{while(true)} loop: inputs are re-sampled nondeterministically (open-world sensor model), outputs and timer/counter state persist across iterations, and YAML safety/integrity properties become \code{\_\_ESBMC\_assert()} statements. \textit{k}-induction proves a property for all scan counts (unbounded), while incremental \gls{bmc} unrolls the loop to return a violating trace. Our modeling layer (\S\ref{sec:impl}) sits in front of this encoding, which is used unchanged}

\label{fig:scan-encoding}
\end{figure}

A \gls{plc} executes a cyclic \emph{scan}: it samples inputs, evaluates the control program once, writes outputs, and repeats, with a watchdog timer aborting the cycle if evaluation exceeds a fixed budget. ESBMC-PLC+~\cite{ESBMCpr5400} encodes this scan cycle as a \code{while(true)} loop in which inputs are re-sampled nondeterministically (an \emph{open-world} sensor model), outputs and timer/counter state are persistent variables, and YAML-specified safety/integrity properties are injected as \code{\_\_ESBMC\_assert()} statements. The resulting model is discharged by the \gls{esbmc}~\cite{gadelha2020,menezes2024} backend: \textit{k}-induction proves a property holds for \emph{all} scan counts (an unbounded guarantee). In contrast, incremental \gls{bmc} unrolls the loop to find a violating trace (Fig.~\ref{fig:scan-encoding}). We use this encoding unchanged and add only a modeling layer in front of it (\S\ref{sec:impl}).

\subsection{Ladder Logic Bombs}
\label{sec:llb-background}
\citet{Govil2017} defines a \gls{llb} by its \emph{trigger} (the activation condition) and its \emph{payload} (the malicious effect), and demonstrates bombs that manipulate actuators, forge sensor/HMI values, and deny control. We reproduce this taxonomy in \S\ref{sec:corpus}. The threat is not hypothetical: the Stuxnet worm sabotaged centrifuge controllers by injecting malicious \gls{plc} logic that remained dormant under inspection and detonated only under a precise process state~\cite{Langner2011stuxnet}; \gls{llb}s generalize this stealth pattern to arbitrary trigger/payload pairs, and their dormancy is exactly what defeats commissioning tests and short simulations.

\subsection{Related Detection Work}
\label{sec:related}
Table~\ref{tab:positioning} groups prior work by approach and positions ESBMC-LLB: on Boolean/integer \gls{ld} it matches dedicated detectors while uniquely adding unbounded bomb-absence proofs and adaptive-trigger robustness, whereas on nonlinear analog control PLC-Defuser leads -- so the two are complementary.

\begin{table*}[htbp]
\centering
\small
\caption{Positioning ESBMC-LLB among \gls{plc} malicious-/unsafe-logic detectors. ``Sound absence proof'' = certifies no bomb (of the modelled class) for \emph{all} scan cycles; ``Trigger synth.''\ = recovers the concrete detonation condition; ``Int.\ trig.\ scale'' = behavior on integer-comparison/counter triggers; ``Adapt.\ robust'' = detects triggers engineered to evade syntactic/CFG triage. \textsuperscript{$\dagger$}ESBMC-LLB handles linear/flat integer arithmetic (SMT-BV); PLC-Defuser's CFG triage localizes triggers structurally, handling nonlinear arithmetic without SMT.}
\label{tab:positioning}
\resizebox{\textwidth}{!}{%
\begin{tabular}{@{}llcccccc@{}}
\toprule
\textbf{Tool} & \textbf{Approach / backend} &
\textbf{Input} & \textbf{Unbound.} & \textbf{Sound} & \textbf{Trigger} &
\textbf{Int.\ trig.} & \textbf{Adapt.} \\
 & & \textbf{format} & \textbf{proof} & \textbf{absence} & \textbf{synth.} &
\textbf{scale} & \textbf{robust} \\
\midrule
\multicolumn{8}{@{}l}{\textbf{Dedicated LLB detectors (closest competitors)}} \\
\textbf{This work} & ESBMC \textit{k}-ind.\ + BMC (SMT-BV) &
  LD+ST (PLCopen) & \textbf{Yes} & \textbf{Yes} & \textbf{Yes} &
  Lin.\ flat\textsuperscript{$\dagger$} & \textbf{Yes} \\
PLC-Defuser~\cite{Rinieri2026} & CFG triage + model checking &
  LD (PLCopen) & No & No & Partial &
  \textbf{Yes\textsuperscript{$\dagger$}} & No \\
Iacobelli et al.~\cite{Iacobelli2024} & Formal verification arch. &
  LD (PLCopen) & No & No & No & OK & No \\  
AttkFinder~\cite{Zhang2021attkfinder} & Information-flow analysis &
  ST/IL & No & No & No & -- & -- \\
\midrule
\multicolumn{8}{@{}l}{\textbf{General PLC formal verification (not LLB-specific)}} \\
TSV~\cite{McLaughlin2014tsv} & Model checking (NuSMV) &
  PLC bytecode & No & No & No & Poor & -- \\ 
SymPLC~\cite{Guo2017symplc} & Symb.\ exec.\ (MATIEC$\to$KLEE) &
  ST (IEC) & No & No & No & Poor & -- \\ 
Arcade.PLC~\cite{Zonouz2014arcade} & Model checking / abstr.\ int. &
  IL/ST & Partial & Partial & No & Poor & -- \\
PLCverif~\cite{LopezMiguel2022} & CBMC / nuXmv &
  SCL/STL & No & No & No & Poor (BDD) & -- \\ 
\midrule
\multicolumn{8}{@{}l}{\textbf{Invariant mining / safety vetting}} \\
VetPLC~\cite{Zhang2019vetplc} & Temporal invariant mining &
  PLC code+traces & No & No & No & -- & -- \\
SAIN~\cite{Abbas2024sain} & State-aware invariants &
  runtime state & No & No & No & -- & -- \\
\midrule
\multicolumn{8}{@{}l}{\textbf{Runtime / anomaly (different layer; not head-to-head)}} \\
Process/ML anomaly (SWaT) & ML on sensor traces &
  network/proc.\ & No & No & No & -- & -- \\
\bottomrule
\end{tabular}%
}
\end{table*}

\textbf{Formal verification / model checking of \gls{plc} code} is the directly comparable line. \citet{Iacobelli2024} (ICPS'24) is the closest competitor: a formal-verification architecture for \gls{llb} detection, evaluated on a public dataset of 30~legitimate and 30~malicious PLCopen \gls{ld} programs in which the bombs are \code{EQ}/\code{LT}/\code{LE}/\code{GE}/\code{GT} comparison-block triggers with coil/assignment payloads. We use \emph{their} dataset for our head-to-head (\S\ref{sec:rq-iacobelli}). The \emph{\gls{tsv}}~\cite{McLaughlin2014tsv} performs bump-in-the-wire model checking of \gls{plc} bytecode against safety properties, but is bounded and not spec-free. \emph{SymPLC}~\cite{Guo2017symplc} applies symbolic execution to IEC~61131-3 via the same MATIEC~\cite{deSousa2014}$\to$C path we use, but with KLEE -- giving coverage-driven test generation, not unbounded proofs. \emph{PLCverif}~\cite{LopezMiguel2022} is based on \gls{cbmc} or \gls{bdd}-bound (nuXmv~\cite{Cavada2014}) and Siemens-specific. None of these provides an unbounded proof of bomb-\emph{absence}, automatic trigger synthesis, or robustness on integer triggers -- the three axes on which ESBMC-LLB leads.

\textbf{General IEC~61131-3 verification:} A broad line formalizes IEC~61131-3 semantics and verifies \gls{plc} code for functional correctness and safety rather than malicious logic: executable \gls{st} semantics~\cite{Wang2023,Lee2024,Lee2025}, formal verification with inductive code synthesis~\cite{Weiss2021}, cooperative verification~\cite{Ukegbu2023a}, ladder-logic fault-impact analysis~\cite{Ebnenasir2023}, monitor-based verification coupling FRET and PLCverif~\cite{Fink2024}, and verification-as-a-service deployments on safety-critical plants~\cite{LopezMiguel2025}. These tools verify a \emph{specification of intended behavior}; none recast \gls{llb} detection as an unbounded bomb-absence proof with trigger synthesis, which is the gap ESBMC-LLB fills.

\textbf{Invariant mining/safety vetting:} \emph{VetPLC}~\cite{Zhang2019vetplc} (CCS'19) mines temporal invariants to vet \gls{plc} code, and \emph{SAIN}~\cite{Abbas2024sain} (USENIX'24) refines state-aware invariants; both target a different problem (discovering specifications/anomalies) and are complementary rather than head-to-head competitors.

\textbf{Runtime attestation and process-level \gls{ml} anomaly detection} (e.g.\ on the SWaT testbed) operate at a different layer (network/sensor traces, not control logic) and detect attacks at execution time rather than certifying the program offline. We position against them qualitatively but do not claim a head-to-head.

\section{Threat Model}
\label{sec:threat}

\begin{figure}[htbp]
\centering
\resizebox{0.55\columnwidth}{!}{%
\begin{tikzpicture}[
  font=\scriptsize,
  node distance=6mm and 9mm,
  box/.style={draw, rounded corners, align=center, inner sep=3pt, minimum height=7mm},
  adv/.style={box, fill=red!8,  draw=red!55},
  sys/.style={box, fill=blue!6, draw=blue!45},
  def/.style={box, fill=teal!12, draw=teal!55},
  ok/.style={box,  fill=green!8, draw=green!50!black},
  bad/.style={box, fill=red!8,  draw=red!60},
  a/.style={-{Latex[length=1.6mm]}, semithick}
]
\node[adv] (adv) {Adversary\\(insider / eng.\ station /\\supply-chain tamper)};
\node[sys, below=of adv] (prog) {PLC program\\\emph{dormant LLB}: trigger\,$\to$\,payload};
\node[sys, right=of prog] (plc) {PLC\\scan cycle};
\node[sys, below=of plc] (plant) {Physical\\process};
\node[def, below=of prog] (tool) {\textbf{ESBMC-LLB}\\FB-body translation\,$+$\,scan-watchdog\\$+$\,output wiring \emph{over} ESBMC-PLC+\\($k$-induction\,$+$\,BMC)};
\node[ok,  below=6mm of tool, xshift=-17mm] (safe) {\textbf{SAFE}\\unbounded\\bomb-absence};
\node[bad, below=6mm of tool, xshift=17mm]  (viol) {\textbf{VIOLATION}\\$+$ detonation\\trigger};

\draw[a, red!55] (adv) -- node[right]{inserts} (prog);
\draw[a] (prog) -- (plc);
\draw[a] (plc) -- node[right]{actuate} (plant);
\draw[a] (plant) to[bend left=35] node[left]{sense} (plc);
\draw[a, teal!60] (prog) -- node[right]{offline check} (tool);
\draw[a] (tool.south) -- (safe.north);
\draw[a] (tool.south) -- (viol.north);
\end{tikzpicture}
}
\caption{Threat model and the \textbf{ESBMC-LLB} verification-in-the-loop defence. An adversary inserts a dormant ladder-logic bomb into the \gls{plc} program driving the physical process; offline, ESBMC-LLB exposes the function-block-resident payload, and either proves bomb-absence across all scan cycles (\textit{k}-induction) or returns a counterexample that \emph{is} the detonation trigger (\gls{bmc}).}

\label{fig:threat}
\end{figure}
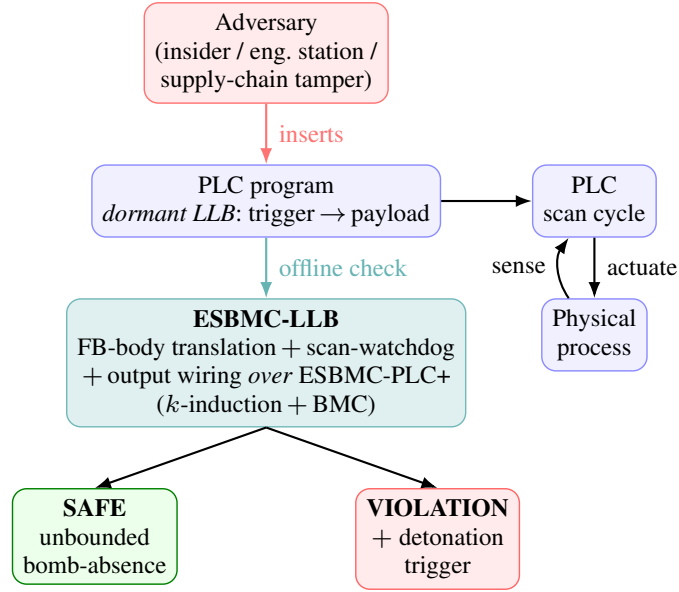

We consider an adversary who targets a \gls{plc}-controlled industrial process by inserting a dormant \gls{llb} into the control program -- a fragment that passes commissioning tests and short simulations because its trigger is engineered to remain inactive under normal operation. The defender's response is an offline, verification-in-the-loop check that runs before the program is deployed to the plant. Figure~\ref{fig:threat} illustrates the resulting architecture; the two items below scope the attacker's capability, and the defender's posture.

\begin{itemize}
    \item \textbf{Adversary:} An attacker with the ability to modify the \gls{plc} program (a malicious insider, a compromised engineering station, or a tampered project/supply chain artifact), but who must keep the bomb dormant during normal operation to evade review and testing.
    
    \item \textbf{Defender:} A verification-in-the-loop process that runs ESBMC-LLB on the deployed program against a set of safety/integrity properties derived from the plant specification. 
\end{itemize}

We assume a single-task scan-cycle model with an open-world (nondeterministic) sensor model; that the supplied properties capture the hazardous or integrity-violating states of interest (or, for availability attacks, that scan non-termination is itself hazardous); and a trusted toolchain (the \gls{esbmc}~backend and the function-block translation). The adversary may obfuscate the trigger arbitrarily, but cannot alter the verifier or the property set.

\section{Approach: \gls{llb} Detection as Property Checking}
\label{sec:approach}

Table~\ref{tab:taxonomy-map} maps the~\citet{Govil2017} taxonomy onto ESBMC-PLC+ property kinds. The key observation is that every payload class has a natural safety/integrity property whose violation is necessary and sufficient for the bomb to have fired.

\begin{table}[t]
\centering
\small
\caption{\citet{Govil2017} \gls{llb} taxonomy mapped to ESBMC-PLC+ properties}
\label{tab:taxonomy-map}
\begin{tabular}{lll}
\toprule
\textbf{Trigger / Payload} & \textbf{ESBMC-PLC+ property} & \textbf{Detection verdict} \\
\midrule
Counter/timer $\to$ actuator manip.    & \texttt{absence}(unsafe combo)        & VIOLATION + trigger \\
Input-pattern $\to$ actuator manip.    & \texttt{mutual\_exclusion}/\texttt{reachability} & VIOLATION + knock \\
Counter/timer $\to$ sensor forgery     & \texttt{invariant}(reported==real)    & VIOLATION + trigger \\
Counter/timer $\to$ denial of control  & \texttt{response}/\texttt{absence}    & VIOLATION + trigger \\
\midrule
Benign baseline (no bomb)              & all of the above                      & SAFE (\textit{k}-induction) \\
\bottomrule
\end{tabular}
\end{table}

The two verification outcomes follow directly from the property-checking reduction: on a benign program, the prover certifies absence; on a bombed program, it returns the concrete detonation condition. Both are products of the same backend, with no change to the underlying engine.

\begin{itemize}
    \item \textbf{Unbounded absence proof:} For a benign program, \textit{k}-induction proves the property holds for all scan counts: a certificate that no bomb of the modeled class is present. 

    \item \textbf{Trigger synthesis:} For a bombed program, incremental \gls{bmc} returns the shortest counterexample; the input valuations and scan index along that trace constitute the detonation trigger.

\end{itemize}

\subsection{Two Detection Mechanisms}
\label{sec:modes}
\begin{figure}[htbp]
\centering
\resizebox{0.9\columnwidth}{!}{%
\begin{tikzpicture}[
  font=\scriptsize,
  >={Latex[length=1.4mm]},
  node distance=5mm and 8mm,
  box/.style={draw, rounded corners, align=center, inner sep=3pt, minimum height=6mm},
  payload/.style={box, fill=red!8,  draw=red!55,        text width=30mm},
  cond/.style={box, fill=blue!6, draw=blue!45, font=\scriptsize\bfseries, text width=34mm},
  mech/.style={box, fill=gray!8, draw=gray!55,          text width=40mm},
  ok/.style={box,  fill=green!8, draw=green!50!black,    text width=42mm},
  fail/.style={box, fill=red!6,  draw=red!50,            text width=35mm},
  a/.style={->, semithick},
  lbl/.style={font=\tiny}
]
\node[payload] (bomb) {\gls{llb} payload hidden in a function-block body};
\node[cond, below=of bomb] (a) {(a)~present in the \gls{ir}?};
\draw[a] (bomb) -- (a);

\node[fail, left=9mm of a] (prior)
  {prior \gls{ld} verifiers: FB body dropped $\Rightarrow$ undetectable~$\times$};
\draw[a, red!55, dashed] (a) -- node[lbl, above]{drop FB} (prior);

\node[ok, below=of a] (trans)
  {FB-body translation (\S\ref{sec:impl}) $\Rightarrow$ payload in \gls{ir}~$\checkmark$};
\draw[a] (a) -- node[lbl, right]{translate} (trans);

\node[cond, below=of trans] (b) {(b)~violates a checkable property?};
\draw[a] (trans) -- (b);

\node[mech, below left=5mm and -7mm of b] (wd)
  {\textbf{scan-watchdog} (non-termination): assert counter $\le$ budget};
\node[mech, below right=5mm and -7mm of b] (ow)
  {\textbf{output wiring} (forgery): forged \gls{fb} output $\to$ \texttt{mutual\_exclusion}};
\draw[a] (b) -| (wd);
\draw[a] (b) -| (ow);

\node[ok, below=16mm of b] (viol)
  {reachable safety violation $\Rightarrow$ \gls{bmc} counterexample \emph{is the trigger}};
\draw[a] (wd) |- (viol);
\draw[a] (ow) |- (viol);
\end{tikzpicture}
}
\caption{When an \gls{llb} is detectable. A payload must be (a)~present in the verifier's \gls{ir} \emph{and} (b)~violate a checkable property. Prior \gls{ld} verifiers drop function-block bodies, so the payload never reaches the \gls{ir} (left); ESBMC-LLB's function-block-body translation (\S\ref{sec:impl}) restores it. Two mechanisms then satisfy (b): a \emph{scan-watchdog} turns a non-terminating payload into a bounded-counter assertion, and \emph{output wiring} makes a forged \gls{fb} output violate a supplied safety property (e.g.\ \texttt{mutual\_exclusion}). Either way, the violation is reachable, and the \gls{bmc} counterexample recovers the trigger}

\label{fig:detect}
\end{figure}
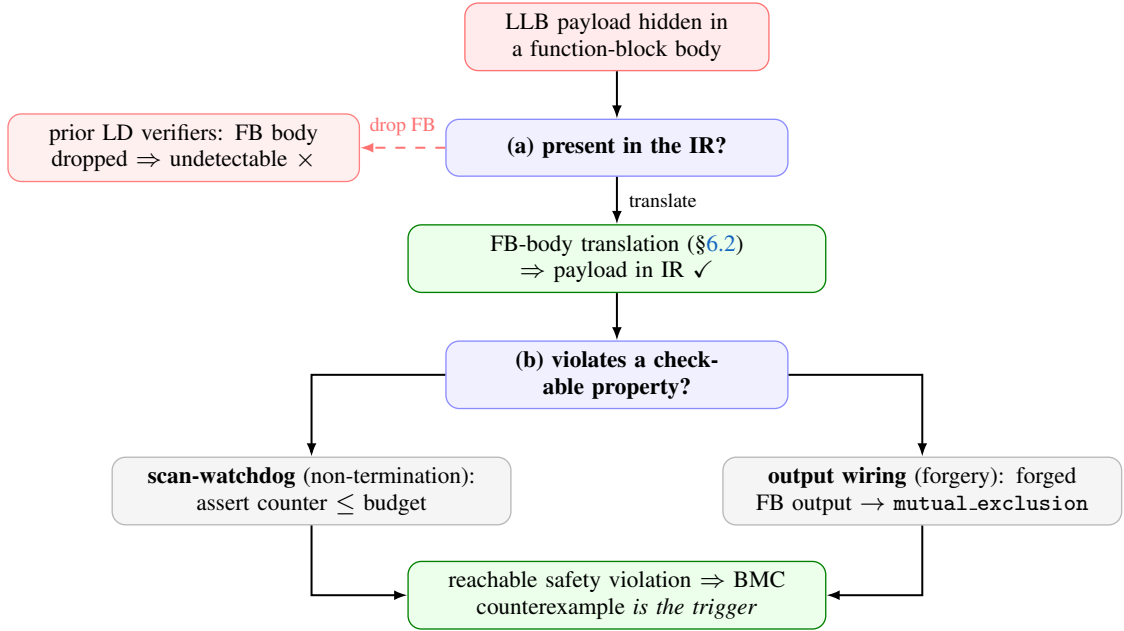

A bomb is detectable only once its payload (a)~is present in the verifier's \gls{ir} and (b)~violates a checkable property. Prior \gls{ld} verifiers fail (a) because \gls{llb} payloads hide inside function-block bodies that are dropped from the \gls{ir}; \S\ref{sec:impl} closes this with function-block-body translation. Given the translated body, two mechanisms cover the two dominant payload families (Figure~\ref{fig:detect}).

\begin{enumerate}
    \item \textbf{Scan-watchdog} (denial-of-control / non-termination): A non-terminating trigger-gated loop is the canonical time-bomb payload. We instrument each rung loop with a bounded iteration counter asserted within budget -- faithful to a real \gls{plc} watchdog timer -- so that a non-terminating payload becomes a reachable safety violation. The counterexample recovers the trigger.

    \item \textbf{Output wiring with safety properties} (actuator/sensor forgery): A value-forgery payload changes a function-block output. By wiring \gls{fb} output pins to the program variables that consume them, a forged output propagates into the program and violates a supplied safety property (e.g.\ \texttt{mutual\_exclusion} on interlocked actuators), again with trigger synthesis.

\end{enumerate}

Both mechanisms are \emph{sound modulo the modeled class}: a payload that neither fails to terminate nor violates a stated property is out of scope, a limitation we make explicit (\S\ref{sec:discussion}). A spec-free \emph{differential} alternative -- checking a suspect program against a trusted reference for per-scan output equivalence -- is a natural extension and is left to future work, where ``trusted reference'' denotes a certified baseline or the pre-update program version.

\section{\gls{llb} Benchmark Corpus}
\label{sec:corpus}

To evaluate ESBMC-LLB across the full range of claims -- third-party detection accuracy, resistance to adaptive adversaries, scalability across the trigger/payload taxonomy, and head-to-head comparison on a real analog benchmark -- we use four corpora assembled in increasing order of independence and difficulty (Table~\ref{tab:corpus}). Together they cover Boolean and integer \gls{ld}, analog process control, and both synthetic and externally authored programs, ensuring that no single result rests on data we designed ourselves.

\begin{table}[t]
\centering
\small
\caption{Evaluation corpora. RQ identifiers refer to the Research Questions defined in \S\ref{sec:rqs}}
\label{tab:corpus}
\begin{tabular}{clcl}
\toprule
\textbf{ID} & \textbf{Corpus} & \textbf{Size} & \textbf{Role} \\
\midrule
C1 & \citet{Iacobelli2024} & 30+30 & third-party detection (RQ2) \\
C2 & Adaptive-adversary variants           & 5     & evasion resistance (RQ4) \\
C3 & Taxonomy corpus (generated)           & 155+155 & Boolean/integer scale (RQ5) \\
C4 & PLC-Defuser SWaT~\cite{Rinieri2026}   & 150+150 & real analog benchmark, 2 versions (RQ6) \\
\bottomrule
\end{tabular}
\end{table}

\begin{enumerate}
    \item \textbf{Third-party dataset (C1):} The public~\citet{Iacobelli2024} dataset -- 30~legitimate and 30~malicious PLCopen \gls{ld} programs modelling a water-treatment process, with the bombs implemented as malicious function-block wrappers (\code{EQ\_0}/\code{LE\_0}/\code{LT\_0}/\code{GE\_0}/\code{GT\_0}/\code{SUB\_0}) -- provides detection results on data we did not create (RQ2). 
    
    \item \textbf{Adaptive-adversary variants (C2):} Five semantically equivalent variants of one dataset bomb whose trigger is hidden from syntactic triage (RQ4). 
    
    \item \textbf{Taxonomy corpus (C3):} A controlled, reproducible generator emits benign/malicious pairs across 6~trigger classes $\times$~2~payload classes $\times$~13~constants (310~Boolean/integer programs) at the scale of the state-of-the-art evaluation (RQ5). 
    
    \item \textbf{PLC-Defuser's real SWaT corpus (C4):} PLC-Defuser's tool and datasets are public~\cite{Rinieri2026}; its headline SWaT corpus (150~legitimate $+$ 150~malicious \emph{analog} programs) is the state-of-the-art's own benchmark. We evaluate two pinned versions -- the archived v1.0.0 release (linear-trigger bombs) and a later development snapshot that adds nonlinear-arithmetic bombs -- as the real analog comparison (RQ6). All programs, properties, and harnesses are in the artifact (\S\ref{sec:artifact}).

\end{enumerate}

\section{Experimental Evaluation}
\label{sec:experiments}
We structure the evaluation around seven research questions that together cover the full chain of claims made in this paper. RQ1 establishes the enabling prerequisite -- that function-block-body translation actually exposes \gls{fb}-resident bombs to the verifier -- without which no downstream detection result is meaningful. RQ2--RQ3 then assess detection accuracy and payload generality on third-party and constructed programs, respectively. RQ4--RQ6 progressively stress the approach: against an adaptive adversary, at the taxonomy scale on Boolean/integer logic, and finally on PLC-Defuser's real analog SWaT benchmark in a direct head-to-head comparison. RQ7 closes with a regression check and a capability-level comparison to the field.

\subsection{Research Questions}
\label{sec:rqs}
\begin{description}[leftmargin=2em]
\item[\textbf{RQ1}] \textbf{Enabling:} Does the function-block-body translation make \gls{llb}s that hide inside \gls{ld} function blocks reachable to the verifier?
\item[\textbf{RQ2}] \textbf{Detection on third-party data:} On the public \citet{Iacobelli2024} dataset (30~legitimate + 30~malicious \gls{ld} programs), what detection rate, false-positive rate, runtime, and trigger recovery does our approach achieve?
\item[\textbf{RQ3}] \textbf{Generality:} Does the approach detect bomb classes beyond non-termination -- specifically value/actuator-forgery payloads with no loop?
\item[\textbf{RQ4}] \textbf{Adaptive adversary:} Does our approach detect triggers engineered to evade a syntactic/\gls{cfg}-based trigger-triage heuristic?
\item[\textbf{RQ5}] \textbf{Scale:} Does detection hold at the scale of the state-of-the-art evaluation on Boolean/integer logic, across the trigger/payload taxonomy?
\item[\textbf{RQ6}] \textbf{Real head-to-head:} On PLC-Defuser's own public SWaT benchmark, can our approach run, and how does it compare?
\item[\textbf{RQ7}] \textbf{Regression and capability:} Are all inherited ESBMC-PLC+ results preserved, and how does our approach compare by capability to the field (Table~\ref{tab:positioning})?
\end{description}

\subsection{Implementation and Setup}
\label{sec:impl}
Our method uses ESBMC-PLC+ as the verification engine \emph{unchanged}; the contribution is a lightweight modeling layer ($\sim$240 lines, plus a small Structured-Text-to-GOTO translator) that prepares a program for ESBMC-PLC+ to detect an \gls{llb}. Three elements close the gap by which \gls{llb}s evaded prior \gls{ld} tools: 

\begin{enumerate}
    \item \textbf{Function-block-body translation:} Each user-defined \texttt{function\-Block} \gls{pou} body is parsed and emitted as native GOTO code (assignments, \code{IF}, \code{WHILE}, comparisons, arithmetic), executed once per scan with nondeterministic inputs;
    
    \item \textbf{A scan-watchdog:} Each rung loop carries a bounded iteration counter asserted within budget, so a (trigger-gated) non-terminating payload becomes a reachable safety violation, faithful to the watchdog timer of a real \gls{plc}; and 
    
    \item \textbf{Output wiring:} An \gls{fb} output pin consumed by a program variable is assigned to that variable. Hence, a forged \gls{fb} output propagates into the program and becomes observable by safety properties.

\end{enumerate}

For the analog benchmark (RQ6), we additionally extend the frontend with REAL types, non-Boolean coils/contacts (numeric--Boolean coercion), and a \gls{st} translator that over-approximates unsupported constructs; we distinguish this \emph{analog-extended configuration} from the \emph{sound Boolean/integer configuration} used for RQ2--RQ5, because the over-approximation trades soundness for coverage (\S\ref{sec:discussion}). All experiments use \gls{esbmc}~v8.3.0 with the Z3 backend on Apple Silicon (aarch64, macOS); \texttt{--k-induction} proves benign programs SAFE and \texttt{--incremental-bmc} searches for violations.

\subsection{RQ1: Enabling -- making \gls{fb}-resident bombs visible}
The \citet{Iacobelli2024} bombs are implemented \emph{inside} function-block bodies (e.g.\ an \code{EQ\_0} wrapper that mimics equality but adds \code{if IN1=12 then while\dots}). Inspecting the GOTO~\gls{ir} of the unmodified ESBMC-PLC+ confirms the root cause: the malicious program declares the \gls{fb} instance, but its body -- and therefore the bomb -- is \emph{absent} from the \gls{ir}; the malicious and benign \gls{ir} are essentially identical so that no property can distinguish them. With function-block-body translation the bomb logic appears in the \gls{ir} (the \code{if IN1=12} guard and the non-terminating loop), and a proof-of-concept on the six \gls{fb} types of the dataset (\code{EQ\_0}/\code{LE\_0}/\code{LT\_0}/\code{GE\_0}/\code{GT\_0}/\code{SUB\_0}) detects all six and recovers each trigger (12 for the comparison FBs, 25 for \code{SUB\_0}), with the benign \gls{fb} proved SAFE. This is the prerequisite for all subsequent results. Figure~\ref{fig:example} shows the mechanism end-to-end on the \code{EQ\_0} bomb.

\begin{figure}[htbp]
\begin{lstlisting}[basicstyle=\ttfamily\scriptsize,frame=single,
  numbers=none,xleftmargin=0.5em]
(* Malicious EQ_0 function block: bomb hidden in the FB body *)
FUNCTION_BLOCK EQ_0
  VAR_INPUT  IN1, IN2 : INT;  END_VAR
  VAR_OUTPUT OUT : BOOL;      END_VAR
  OUT := (IN1 = IN2);            (* benign equality   *)
  IF IN1 = 12 THEN               (* dormant trigger   *)
    WHILE TRUE DO ; END_WHILE;   (* denial-of-control *)
  END_IF;
END_FUNCTION_BLOCK

// ESBMC-LLB: FB-body translation + scan-watchdog (GOTO IR)
OUT = (IN1 == IN2);
if (IN1 == 12)
  for (w = 0; ; ++w)
    __ESBMC_assert(w < WD_BUDGET);  // reachable iff IN1 == 12

// incremental-bmc  -> counterexample  IN1 = 12   (detonation trigger)
// k-induction (benign EQ, no IF)      -> SAFE     (bomb-absence)
\end{lstlisting}
\caption{Enabling mechanism on the \code{EQ\_0} bomb. The unmodified \gls{ld} frontend drops the function-block body, so the malicious and benign programs share an identical \gls{ir}. ESBMC-LLB translates the body and instruments the rung loop with a scan-watchdog assertion; the non-terminating payload becomes a reachable safety violation whose \gls{bmc} counterexample (\code{IN1}\,$=$\,12) \emph{is} the trigger. At the same time, \textit{k}-induction certifies that the block is benign.} 
\label{fig:example}
\end{figure}

\subsection{RQ2: Detection on the \citet{Iacobelli2024} dataset}
\label{sec:rq-iacobelli}
We run ESBMC-LLB natively (no MATIEC -- these are PLCopen \gls{ld}) on all 60~programs of~\cite{Iacobelli2024}. Table~\ref{tab:iacobelli} reports the confusion matrix. \textbf{All 30~malicious programs are detected} (recall \SI{100}{\percent}, zero false negatives) and \textbf{29 of 30~benign programs are proved SAFE with zero false positives}; the single non-SAFE benign program is \emph{inconclusive} (\texttt{UNKNOWN}), not a false alarm -- it is genuinely safe but its integer state defeats \textit{k}-induction's inductive step. Detection is sub-second per program. The counterexample recovers the detonation trigger in every detected case (constants 5, 12, 25, 30, 46, 54 across families). This matches the ``very high accuracy and precision'' reported by~\cite{Iacobelli2024} on the same data, while also providing unbounded proofs of the absence of bombs and automatic trigger synthesis. These figures are for the sound Boolean/integer configuration; the analog-extended configuration of RQ6, when applied to this same set, introduces one false positive (\S\ref{sec:discussion}), which is why we keep the two configurations distinct.

\begin{table}[t]
\centering\small
\caption{Detection on the \citet{Iacobelli2024} dataset (60 programs)}
\label{tab:iacobelli}
\begin{tabular}{lcccc}
\toprule
 & \textbf{detected} & \textbf{missed} & \textbf{inconclusive} & \textbf{rate} \\
\midrule
Malicious (30) & \textbf{30} (TP) & 0 (FN) & 0 & recall \SI{100}{\percent} \\
Benign (30)    & 0 (FP) & \textbf{29} (TN) & 1 & FP \SI{0}{\percent} \\
\bottomrule
\end{tabular}
\end{table}

\subsection{RQ3: Generality beyond non-termination}
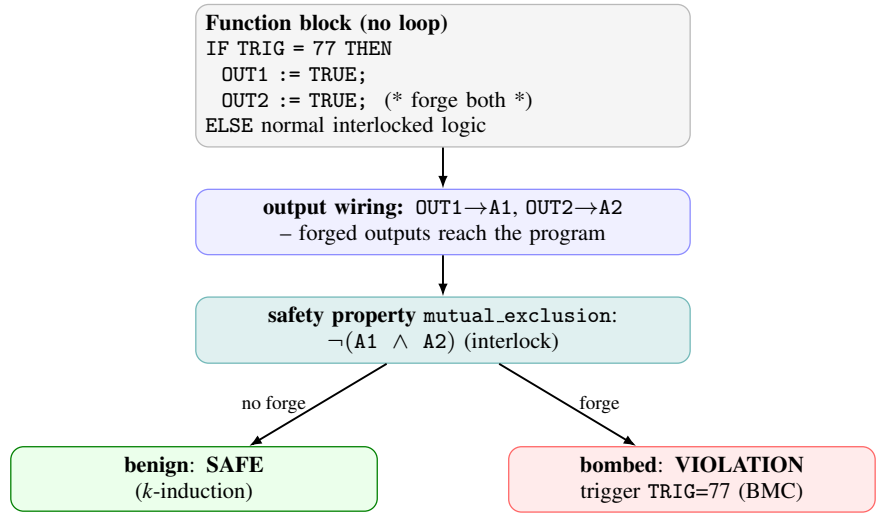
\begin{figure}[htbp]
\centering
\resizebox{0.7\columnwidth}{!}{%
\begin{tikzpicture}[
  font=\scriptsize,
  >={Latex[length=1.4mm]},
  node distance=4.5mm,
  box/.style={draw, rounded corners, align=center, inner sep=3pt, minimum height=6mm},
  code/.style={box, fill=gray!8,  draw=gray!55, align=left,
               font=\ttfamily\scriptsize, text width=52mm},
  proc/.style={box, fill=blue!6,  draw=blue!45, text width=52mm},
  prop/.style={box, fill=teal!12, draw=teal!55, text width=52mm},
  ok/.style={box,   fill=green!8, draw=green!50!black, text width=38mm},
  bad/.style={box,  fill=red!8,   draw=red!60,         text width=38mm},
  a/.style={->, semithick},
  lbl/.style={font=\tiny}
]
\node[code] (fb)
{\textrm{\textbf{Function block (no loop)}}\\
IF TRIG = 77 THEN\\
\ \ OUT1 := TRUE;\\
\ \ OUT2 := TRUE;\ \ \textrm{(* forge both *)}\\
ELSE \textrm{normal interlocked logic}};

\node[proc, below=of fb] (wire)
{\textbf{output wiring:} \code{OUT1}$\to$\code{A1},\ \code{OUT2}$\to$\code{A2} -- forged outputs reach the program};
\draw[a] (fb) -- (wire);

\node[prop, below=of wire] (prop)
{\textbf{safety property} \texttt{mutual\_exclusion}: $\lnot(\code{A1}\wedge\code{A2})$ (interlock)};
\draw[a] (wire) -- (prop);

\node[ok,  below left=9mm and -2cm of prop] (safe)
{\textbf{benign}: \textbf{SAFE}\\ (\textit{k}-induction)};
\node[bad, below right=9mm and -2cm of prop] (viol)
{\textbf{bombed}: \textbf{VIOLATION}\\ trigger \code{TRIG}=77 (\gls{bmc})};
\draw[a] (prop) -- node[lbl, left]{no forge} (safe);
\draw[a] (prop) -- node[lbl, right]{forge} (viol);
\end{tikzpicture}
}
\caption{RQ3 generality: a no-loop \emph{actuator-forgery} bomb, detected via a safety property rather than the scan-watchdog. A function block drives two interlocked actuators; under the trigger \code{TRIG}=77, it forges both \code{OUT1} and \code{OUT2} to \code{TRUE}, breaking the interlock. Output wiring propagates the forged outputs to the program variables \code{A1}, \code{A2}, so the \texttt{mutual\_exclusion} property $\lnot(\code{A1}\wedge\code{A2})$ becomes checkable: the benign program is proved \textbf{SAFE} by \textit{k}-induction, while the bombed variant is reported \textbf{VIOLATION} with the trigger recovered. This covers the actuator-forgery \gls{llb} class through a mechanism orthogonal to the watchdog}

\label{fig:forgery}
\end{figure}

The Iacobelli payloads are all non-termination (caught by the scan-watchdog). To test generality, we construct a \emph{value/actuator-forgery} bomb with \emph{no loop}: a function block drives two interlocked actuators and, under a trigger value, forges both \code{TRUE}, breaking the interlock. With output wiring, the forged outputs reach the program; the benign program is proved SAFE for the \texttt{mutual\_exclusion} property, while the bombed variant is reported \texttt{VIOLATION} with the trigger (\code{TRIG}=77) recovered -- detection via a \emph{safety property}, a mechanism orthogonal to the watchdog. Our approach thus covers two distinct \gls{llb} classes (denial-of-control and actuator forgery) via two complementary mechanisms, each involving trigger synthesis (Fig.~\ref{fig:forgery}).

\subsection{RQ4: Adaptive adversary}
\label{sec:rq-adaptive}
A detector that \emph{triages candidate triggers by syntactic/structural pattern} before checking (the \gls{cfg} step of PLC-Defuser~\cite{Rinieri2026} and signature scanners) can be evaded by hiding the trigger while preserving behavior. We construct five semantically equivalent variants of the \code{EQ\_0} bomb (Table~\ref{tab:adaptive}) and compare our approach against a transparent \emph{\gls{cfg}-triage proxy} that flags the canonical signature -- an input equality-compared to a constant literal guarding a payload. The proxy flags only the naive baseline (1/5); \textbf{our approach detects all five} (5/5), because it evaluates the actual semantics rather than a pattern. For the opaque-arithmetic variants it even finds \emph{bitvector-overflow} detonating inputs (e.g.\ \code{IN1}=1{,}073{,}840{,}131 with $\code{IN1}^2\equiv144\ (\mathrm{mod}\ 2^{32})$) that the attacker's intended $\pm12$ arithmetic did not anticipate. The proxy is a conservative model of syntactic triage, not PLC-Defuser itself; the defensible claim is the general one: \emph{pattern-based triage is evadable, full semantic checking is not.}

\begin{table}[t]
\centering\small
\caption{Adaptive-adversary variants of the \code{EQ\_0} bomb}
\label{tab:adaptive}
\begin{tabular}{lcc}
\toprule
\textbf{Variant (trigger hidden as)} & \textbf{CFG-triage proxy} & \textbf{ESBMC-LLB} \\
\midrule
B0 literal \code{IN1=12}              & FLAGGED & VIOLATION \\
E1 computed \code{k:=5+7; IN1=k}      & missed  & VIOLATION \\
E2 square \code{IN1*IN1=144}          & missed  & VIOLATION \\
E3 factored \code{(IN1-12)\textasciicircum2=0} & missed & VIOLATION \\
E4 temporal (multi-scan accumulator)  & missed  & VIOLATION \\
\bottomrule
\end{tabular}
\end{table}

\subsection{RQ5: Scale on Boolean/integer logic}
We evaluate at the scale of the state-of-the-art evaluation on a controlled, reproducible bomb-injection corpus of Boolean/integer programs spanning \textbf{6 trigger classes $\times$ 2 payload classes $\times$ 13 constants $=$ 155 malicious $+$ 155 benign $=$ 310 programs}. In this \emph{sound (Boolean/integer) configuration}, our approach achieves \textbf{\SI{100}{\percent} recall (155/155) and zero false positives}, with median detection time \SI{70}{\milli\second} (max \SI{305}{\milli\second}), uniform across all trigger classes (including the hidden ones of RQ4) and both payload classes. This is a controlled scalability/breadth result; the third-party validation is RQ2, and the real-benchmark head-to-head is RQ6.

\subsection{RQ6: The real analog benchmark (PLC-Defuser SWaT)}
\label{sec:rq-swat}
PLC-Defuser's tool and datasets are publicly available~\cite {Rinieri2026}. We evaluate on its headline SWaT corpus (150~legitimate $+$ 150~malicious programs modeling the Secure Water Treatment plant) in \emph{two versions} that form two difficulty tiers: the archived release \textbf{v1.0.0}~\cite{Rinieri2026artefact}, whose bombs are linear-trigger non-termination payloads, and a later \textbf{development snapshot} (commit-pinned in our artifact) that adds \emph{nonlinear}-arithmetic non-termination bombs (e.g.\ \code{i := i*i}). This is, to our knowledge, the \emph{first} evaluation of a semantic model checker on this benchmark, and we report both tiers to delineate exactly where the approach holds and where it breaks.

\textbf{Parsing (the enabling result):} SWaT is \emph{analog} process control: programs drive non-Boolean (INT/REAL) coils, use REAL arithmetic, and call timer function blocks. The unmodified \gls{ld} frontend -- Boolean-ladder oriented -- rejects $\approx$\SI{95}{\percent} of either version at parse time. Our analog modeling extension (REAL types, non-Boolean coils, a tolerant \gls{st} translator) makes \textbf{all 300~programs parse, in both versions, for the first time}.

\textbf{Detection on v1.0.0 (linear triggers):} In the analog-extended configuration, we detect \textbf{149/150} malicious programs (Table~\ref{tab:swat}), with \textbf{0/150 false positives}, each in under a second; incremental \gls{bmc} recovers the detonation trigger exactly for the input- and timer-gated cases (e.g.\ \code{IN1}=20). The fully \emph{sound} configuration detects \textbf{75/150} -- it covers every input-triggered bomb but conservatively drops the exception-wrapped (\code{\_\_TRY/\_\_CATCH}) and some timer payloads it cannot faithfully translate, rather than over-approximate them. These bombs are linear-trigger non-termination payloads, so \SI{99}{\percent} detection demonstrates that the enabling pipeline -- analog parsing, function-block-body translation, and the scan-watchdog -- works end-to-end on a \emph{real} analog benchmark, not that it beats a hard problem.

\textbf{Detection on the development snapshot (nonlinear triggers):} On the later corpus, detection drops to \textbf{73/150} (Table~\ref{tab:swat}): we still detect the entire timer category (\textbf{50/50}), but the misses are non-termination bombs whose loop bodies use \emph{nonlinear} arithmetic (\code{i := i*i}), which forces the \gls{smt} solver to reason about $i^{2^k}$ over the loop unrolling and times out. We confirmed this is fundamental, not a tuning artifact: it persists across solvers (Z3, Boolector), unwind depths, and formula-shrinking encodings. PLC-Defuser's \gls{cfg} triage avoids this by structurally localizing the trigger, without solving the nonlinear formula.

\begin{table}[t]
\centering\small
\caption{ESBMC-LLB on PLC-Defuser's SWaT benchmark, two corpus versions (150 malicious, 150 legitimate each)}
\label{tab:swat}
\begin{tabular}{lcc}
\toprule
 & \textbf{v1.0.0 (linear)} & \textbf{dev snapshot (nonlinear)} \\
\textbf{Bomb category} & \textbf{detected} & \textbf{detected} \\
\midrule
timer            & 50/50 & 50/50 \\
particular\_input & 50/50 & 12/50 \\
fault\_code      & 49/50 & 11/50 \\
\midrule
\textbf{Total (analog-extended)} & \textbf{149/150 (\SI{99}{\percent})} & \textbf{73/150 (\SI{49}{\percent})} \\
Sound configuration              & 75/150 & -- \\
Legitimate false positives       & 0/150  & 0/150 \\
\bottomrule
\end{tabular}
\end{table}

\textbf{Honest reading:} On linear-trigger analog non-termination ESBMC-LLB detects \SI{99}{\percent} with zero false positives and full trigger synthesis; on nonlinear-arithmetic non-termination it reaches only \SI{49}{\percent}, clearly behind a \gls{cfg}-triage detector on its own harder corpus. The two paradigms are \emph{complementary} (\S\ref{sec:discussion}): semantic model checking adds unbounded bomb-absence proofs and adaptive-trigger robustness and handles linear analog control. At the same time, \gls{cfg}-triage leads to nonlinear analog non-termination. We do \emph{not} claim parity on the nonlinear tier.

\subsection{RQ7: Regression and capability comparison}
Re-running the full ESBMC-PLC+ artifact benchmark suite under the extended frontend reproduces all inherited verdicts (\textbf{13/13}); the \gls{fb}-body translation, watchdog, and wiring introduce no regression on programs without user FBs (untranslatable \gls{fb} bodies fall back to a sound no-op). Table~\ref{tab:positioning} positions our approach against the field: on Boolean/integer \gls{ld} it matches dedicated detectors and adds unbounded bomb-absence proofs and adaptive-trigger robustness; on nonlinear analog control, it is currently behind PLC-Defuser (RQ6). SymPLC shares our \gls{st} translation path but yields coverage rather than proofs; \gls{tsv} and PLCverif are bounded; VetPLC addresses a complementary problem.

\section{Discussion and Threats to Validity}
\label{sec:discussion}
\textbf{Detection completeness:} ESBMC-LLB is sound for the modeled bomb classes (non-termination and property-violating forgery) but is \emph{complete only modulo the supplied specification and these mechanisms}: a payload that always terminates and violates no stated safety property is out of scope. This is the standard limitation of specification-based detection; the watchdog mechanism narrows it by capturing availability attacks without a bespoke property.

\textbf{Soundness of the over-approximation (measured false positive):} The sound Boolean/integer configuration (RQ2--RQ5) produces zero false positives. The analog-extended configuration's tolerant translator over-approximates unsupported constructs (function results, member access) nondeterministically; this is the price of parsing analog programs, and it is \emph{not} uniformly sound: it introduced \textbf{one false positive} on the Iacobelli benign set (29/30 SAFE, 1~SAFE-program flagged) that the sound configuration does not exhibit. We therefore report the analog results as an exploratory frontier, with the strong soundness claims confined to the Boolean/integer configuration.

\textbf{Nonlinear arithmetic (the limit on RQ6):} The archived v1.0.0 SWaT bombs are linear-trigger non-termination payloads, which we detect at \SI{99}{\percent} (RQ6). The \emph{development} version of the corpus adds non-termination loops with nonlinear updates (\code{i := i*i}); detecting these via the watchdog requires the \gls{smt} solver to reason about $i^{2^k}$ across the unrolling, which times out, dropping detection to \SI{49}{\percent}. We verified this is not a tuning artifact -- it persists across solvers, unwind depths, and formula-shrinking encodings -- and reflects a genuine difference from \gls{cfg}-triage, which localizes the trigger without solving the nonlinear formula. Loop summarization/acceleration or a \gls{cfg}-guided hybrid is the natural remedy and is future work.

\textbf{Corpus realism:} The scale corpus (C3) is synthetic and varies along the trigger/payload taxonomy from a shared host skeleton. Third-party validity rests on the Iacobelli set (C1) and on PLC-Defuser's real SWaT corpus (RQ6), which we did not author.

\textbf{Comparison to PLC-Defuser:} PLC-Defuser's tool and datasets are public, so RQ6 runs on its real SWaT benchmark in both corpus versions. For the archived v1.0.0 release, we detect $149/150$ (\SI{99}{\percent}) with zero false positives; for the later nonlinear-bomb version, detection drops to \SI{49}{\percent}, with PLC-Defuser's \gls{cfg} triage ahead. The adaptive-adversary comparison (RQ4) uses a transparent \gls{cfg}-triage \emph{model}, not the tool itself; reproducing it with the released tool is left to future work. We therefore claim \emph{complementary} strengths, not superiority: parity on Boolean/integer logic and linear analog non-termination, plus unbounded proofs and adaptive robustness, against PLC-Defuser's lead on \emph{nonlinear} analog detection.

\textbf{Verification completeness:} \textit{k}-induction is incomplete; one benign program (C1) is reported \texttt{UNKNOWN} rather than SAFE. Stronger invariant inference or a larger~$k$ would likely resolve it.

\textbf{Measurement validity:} All runtimes are from a single Apple-Silicon host (\gls{esbmc}~v8.3.0, Z3); absolute timings will vary with hardware, solver, and solver version. The \emph{verdicts} (SAFE / VIOLATION / UNKNOWN) and the recovered triggers are deterministic and hardware-independent, and the sub-second medians leave a wide margin below any practical scan-time budget, so we do not expect the qualitative conclusions to depend on the measurement platform. A multi-platform timing study is left to future work.

\section{Conclusion}
\label{sec:conclusion}
We presented a method that turns the ESBMC-PLC+ safety verifier into a \gls{llb} detector. The key observation is that \gls{llb}s in real datasets hide inside \gls{ld} function-block bodies that prior verifiers drop from the intermediate representation; a $\sim$240-line function-block-body translation in the open-source \gls{esbmc} frontend -- with no backend change -- makes them visible. A scan-watchdog and function-block-output wiring then expose, respectively, denial-of-control and actuator-forgery payloads as checkable safety violations, and the \gls{bmc} counterexample recovers each detonation trigger. At the same time, \textit{k}-induction certifies the absence of bombs unboundedly. On the public \citet{Iacobelli2024} dataset, our approach detects all 30~malicious programs and recovers every trigger; on a 310-program Boolean/integer taxonomy corpus, it sustains \SI{100}{\percent} recall and zero false positives at a median \SI{70}{\milli\second}; and, because it reasons over semantics rather than syntactic patterns, it detects adaptive triggers that a control-flow-graph triage misses. On PLC-Defuser's real analog SWaT benchmark, we close the parsing gap that had prevented any semantic model checker from running on the corpus: on the archived v1.0.0 release, we detect $149/150$ bombs (\SI{99}{\percent}) with zero false positives, while on a later corpus version that adds \emph{nonlinear} non-termination bombs detection drops to \SI{49}{\percent} as the \gls{smt} backend times out -- the boundary where a \gls{cfg}-triage detector retains the edge.

We therefore position semantic model checking and \gls{cfg}-triage as \emph{complementary}. For Boolean/integer ladder logic and linear analog non-termination, our approach matches that of dedicated detectors. It adds two guarantees their architecture lacks -- unbounded bomb-absence proofs and adaptive-adversary robustness -- while \gls{cfg}-triage remains superior on \emph{nonlinear} analog non-termination. Bridging the two (loop summarization, or a \gls{cfg}-guided semantic check) is a promising direction toward a detector that is both complete and sound across the full IEC~61131-3 spectrum.

\subsection*{Artefact Availability}
\label{sec:artifact}
The implementation (a branch of the open-source \gls{esbmc} \gls{ld} frontend adding function-block-body translation, the scan-watchdog, and output wiring), the three evaluation corpora (the third-party Iacobelli et al.\ dataset, the adaptive-adversary variants, and the taxonomy-corpus generator), all YAML property files, and the run harnesses accompany this paper. A single binary reproduces both the sound Boolean/integer configuration and the analog-extended configuration via the \code{--ld-sound-mode} flag (with \code{--ld-scan-watchdog}/\code{--ld-scan-budget} controlling the scan-watchdog); the SWaT corpus is pinned to two fixed versions -- the archived Zenodo v1.0.0 release and a commit-pinned development snapshot -- so both RQ6 tiers reproduce. \code{run\_all.sh} runs all experiments and \code{expected/EXPECTED.md} lists the reference results.

For review, the artifact is provided through an anonymized repository linked from the submission; the permanent Zenodo archive and the underlying ESBMC-PLC+ artifact citation are withheld here to preserve anonymity and will be restored in the camera-ready version.


\begin{acks}
    The authors would like to express their gratitude to the Department of Computer Science at the University of Manchester (UoM) and the Systems and Software Security (S3) Research Group for their invaluable support, collaborative environment, and access to cutting-edge resources, which were instrumental in the success of this research. We conducted this work with partial funding from the Engineering and Physical Sciences Research Council (EPSRC) grants EP/T026995/1, EP/V000497/1, EP/X037290/1, and the Soteria project, awarded by the UK Research and Innovation under the Digital Security by Design (DSbD) Programme.
\end{acks}

\bibliographystyle{plainnat}
\bibliography{0.sample-base}

@String{Computer = "{IEEE} Computer" }

@String{Springer = "Springer-Verlag" }

@article{Langner2011stuxnet,
title = {Stuxnet: Dissecting a Cyberwarfare Weapon},
volume = {9},
ISSN = {1540-7993},
DOI = {10.1109/msp.2011.67},
number = {3},
journal = {IEEE Security \& Privacy Magazine},
publisher = {Institute of Electrical and Electronics Engineers (IEEE)},
author = {Langner, Ralph},
year = {2011},
month = May,
pages = {49–51}
}

@inproceedings{McLaughlin2014tsv,
title = {A Trusted Safety Verifier for Process Controller Code},
booktitle = {Proceedings 2014 Network and Distributed System Security Symposium ({NDSS}'14)},
publisher = {Internet Society},
author = {McLaughlin, Stephen and Zonouz, Saman and Pohly, Devin and McDaniel, Patrick},
year = {2014},
month = feb,
address = {San Diego, CA, USA},
numpages = {14},
doi = {10.14722/ndss.2014.23043},
series = {NDSS 2014},
}

@inproceedings{Guo2017symplc,
series = {ESEC/FSE'17},
title = {Symbolic Execution of Programmable Logic Controller Code},
DOI = {10.1145/3106237.3106245},
booktitle = {Proceedings of the 2017 11th Joint Meeting on Foundations of Software Engineering},
publisher = {ACM},
author = {Guo, Shengjian and Wu, Meng and Wang, Chao},
year = {2017},
month = sep,
address = {Paderborn, Germany},
pages = {326--336},
collection = {ESEC/FSE'17},
}

@inproceedings{Zhang2019vetplc,
title = {Towards Automated Safety Vetting of PLC Code in Real-World Plants},
DOI = {10.1109/sp.2019.00034},
booktitle = {2019 IEEE Symposium on Security and Privacy (SP)},
publisher = {IEEE},
author = {Zhang, Mu and Chen, Chien-Ying and Kao, Bin-Chou and Qamsane, Yassine and Shao, Yuru and Lin, Yikai and Shi, Elaine and Mohan, Sibin and Barton, Kira and Moyne, James and Mao, Z. Morley},
address = {San Francisco, CA, USA},
year = {2019},
month = May,
pages = {522–538}
}

@inproceedings{Abbas2024sain,
author = {Abbas, Syed Ghazanfar and Ozmen, Muslum Ozgur and Alsaheel, Abdulellah and Khan, Arslan and Celik, Z. Berkay and Xu, Dongyan},
title = {{SAIN}: Improving {ICS} Attack Detection Sensitivity via State-Aware Invariants},
booktitle = {Proceedings of the 33rd USENIX Security Symposium (USENIX Security 2024)},
year = {2024},
address = {Philadelphia, PA},
pages = {6597--6613},
month = aug,
publisher = {USENIX Association},
url = {https://dl.acm.org/doi/10.5555/3698900.3699269},
}

@inproceedings{Weiss2021,
title = {Towards Establishing Formal Verification and Inductive Code Synthesis in the {PLC} Domain},
DOI = {10.1109/INDIN45523.2021.9557423},
booktitle = {2021 IEEE 19th International Conference on Industrial Informatics (INDIN)},
publisher = {IEEE},
author = {Weis, Matthias and Marks, Philipp and Maschler, Benjamin and White, Dustin and Kesseli, Pascal and Weyrich, Michael},
year = {2021},
month = jul,
address = {Palma de Mallorca, Spain},
pages = {1--8},
numpages = {8},
}

@inproceedings{LopezMiguel2022,
author = {Tournier, Jean-Charles and Fern\'{a}ndez Adiego, Borja and Lopez-Miguel, Ignacio D.},
title = {{PLCverif}: Status of a Formal Verification Tool for Programmable Logic Controller},
booktitle = {Proceedings of the 18th International Conference on Accelerator and Large Experimental Physics Control Systems ({ICALEPCS}'21)},
pages = {MOPV042},
publisher = {JACoW Publishing},
address = {Shanghai, China},
month = oct,
year = {2021},
doi = {10.18429/JACoW-ICALEPCS2021-MOPV042},
isbn = {978-3-95450-221-9},
}

@inproceedings{Ukegbu2023a,
series = {CPS-IoT Week '23},
title = {Cooperative Verification of {PLC} Programs Using {CoVeriTeam}: Towards a Reliable and Secure Industrial Control Systems},
DOI = {10.1145/3576914.3587490},
booktitle = {Proceedings of Cyber-Physical Systems and Internet of Things Week 2023},
publisher = {ACM},
author = {Ukegbu, Chibuzo and Mehrpouyan, Hoda},
year = {2023},
month = may,
address = {San Antonio, TX, USA},
pages = {37--42},
numpages = {6},
collection = {CPS-IoT Week '23},
}

@techreport{Ebnenasir2023,
author = {Ebnenasir, Ali},
title = {Formalizing Ladder Logic Programs and Timing Charts for Fault Impact Analysis and Verification of Fault Tolerance},
institution = {Michigan Technological University, Department of Computer Science},
year = {2023},
number = {CS-TR-23-01},
url = {https://www.mtu.edu/cs/research/papers/pdfs/formalizing-ladder-logic-ali-ebnenasir-tech-rpt-010623-rev.pdf}
}

@article{Wang2023,
title = {K-ST: A Formal Executable Semantics of the Structured Text Language for PLCs},
volume = {49},
ISSN = {2326-3881},
DOI = {10.1109/tse.2023.3315292},
number = {10},
journal = {IEEE Transactions on Software Engineering},
publisher = {Institute of Electrical and Electronics Engineers (IEEE)},
author = {Wang, Kun and Wang, Jingyi and Poskitt, Christopher M. and Chen, Xiangxiang and Sun, Jun and Cheng, Peng},
year = {2023},
month = Oct,
pages = {4796–4813}
}

@inbook{Lee2024,
title = {Formal Semantics and Analysis of Multitask {PLC} {ST} Programs with Preemption},
ISBN = {9783031711626},
ISSN = {1611-3349},
DOI = {10.1007/978-3-031-71162-6_22},
booktitle = {Formal Methods -- 26th International Symposium, {FM} 2024},
publisher = {Springer Nature Switzerland},
author = {Lee, Jaeseo and Bae, Kyungmin},
year = {2024},
month = sep,
address = {Milan, Italy},
pages = {425--442},
}

@inproceedings{Iacobelli2024,
title = {Detection of Ladder Logic Bombs in {PLC} Control Programs: An Architecture Based on Formal Verification},
DOI = {10.1109/ICPS59941.2024.10639995},
booktitle = {2024 IEEE 7th International Conference on Industrial Cyber-Physical Systems (ICPS)},
publisher = {IEEE},
author = {Iacobelli, Antonio and Rinieri, Lorenzo and Melis, Andrea and Sadi, Amir Al and Prandini, Marco and Callegati, Franco},
year = {2024},
month = may,
address = {St.\ Louis, MO, USA},
pages = {1--7},
}

@inbook{Fink2024,
title = {Verifying PLC Programs via Monitors: Extending the Integration of FRET and PLCverif},
ISBN = {9783031606984},
ISSN = {1611-3349},
DOI = {10.1007/978-3-031-60698-4_26},
booktitle = {NASA Formal Methods},
publisher = {Springer Nature Switzerland},
address = {Moffett Field, CA, USA},
author = {Fink, Xaver and Mavridou, Anastasia and Katis, Andreas and Adiego, Borja Fernández},
year = {2024},
pages = {427–435}
}

@misc{Bruttomesso2024,
author = {Bruttomesso, Roberto and {Di Pinto}, Alessandro and Carullo, Moreno and Carcano, Andrea},
title = {Method for automatic translation of ladder logic to a {SMT}-based model checker in a network},
howpublished = {US Patent 11,906,943. Assignee: Nozomi Networks {SAGL}},
year = {2024},
note = {Filed: 2021-08-12. Granted: 2024-02-20.},
}

@inbook{LopezMiguel2025,
title = {Formal Verification of {PLCs} as a Service: A {CERN-GSI} Safety-Critical Case Study},
ISBN = {9783031937064},
ISSN = {1611-3349},
DOI = {10.1007/978-3-031-93706-4_13},
booktitle = {NASA Formal Methods -- 17th International Symposium, {NFM} 2025},
publisher = {Springer Nature Switzerland},
author = {Lopez-Miguel, Ignacio D. and Adiego, Borja Fern\'{a}ndez and Salinas, Matias and Betz, Christine},
year = {2025},
month = jun,
address = {Williamsburg, VA, USA},
pages = {227--235},
}

@inbook{Lee2025,
title = {Formal Analysis of Networked {PLC} Controllers Interacting with Physical Environments},
ISBN = {9783032071064},
ISSN = {1611-3349},
DOI = {10.1007/978-3-032-07106-4_14},
booktitle = {Static Analysis -- 32nd International Symposium, {SAS} 2025},
publisher = {Springer Nature Switzerland},
author = {Lee, Jaeseo and Bae, Kyungmin},
year = {2025},
month = oct,
address = {Singapore, Singapore},
pages = {328--356},
}

@article{gadelha2020,
title = {{ESBMC 6.1: Automated Test Case Generation Using Bounded Model Checking}},
volume = {23},
doi = {10.1007/s10009-020-00571-2},
number = {6},
journal = {International Journal on Software Tools for Technology Transfer},
publisher = {Springer Science and Business Media LLC},
author = {Gadelha, Mikhail R. and Menezes, Rafael S. and Cordeiro, Lucas C.},
year = {2020},
month = may,
pages = {857--861}
}

@inproceedings{menezes2024,
title = {{ESBMC v7.4: Harnessing the Power of Intervals: (Competition Contribution)}},
doi = {10.1007/978-3-031-57256-2_24},
booktitle = {Tools and Algorithms for the Construction and Analysis of Systems (TACAS 2024)},
series = {Lecture Notes in Computer Science},
volume = {14572},
publisher = {Springer Nature Switzerland},
author = {Menezes, Rafael S{\'a} and Aldughaim, Mohannad and Farias, Bruno and Li, Xianzhiyu and Manino, Edoardo and Shmarov, Fedor and Song, Kunjian and Brau{\ss}e, Franz and Gadelha, Mikhail R. and Tihanyi, Norbert and Korovin, Konstantin and Cordeiro, Lucas C.},
year = {2024},
pages = {376--380},
address = {Luxembourg City, Luxembourg},
}

@inproceedings{Cavada2014,
author = {Cavada, Roberto and Cimatti, Alessandro and Dorigatti, Marco and Griggio, Alberto and Mariotti, Alessandro and Micheli, Andrea and Mover, Sergio and Roveri, Marco and Tonetta, Stefano},
title = {The {nuXmv} Symbolic Model Checker},
booktitle = {Computer Aided Verification (CAV 2014)},
series = {Lecture Notes in Computer Science},
volume = {8559},
pages = {334--345},
year = {2014},
publisher = {Springer},
doi = {10.1007/978-3-319-08867-9_22},
address = {Vienna, Austria},
}

@misc{ESBMCpr5400,
author = {Dantas, Pierre and Cordeiro, Lucas C. and {Silva J{\'u}nior}, W.~S.},
title = {{ESBMC-PLC+}: Unified {IEC}~61131-3 Formal Verification Framework with {ST} Frontend and Graphical Function Block Support (Pull Request \#5427)},
year = {2026},
howpublished = {GitHub Pull Request \#5427, \texttt{esbmc/esbmc}},
url = {https://github.com/esbmc/esbmc/pull/5427},
note = {Source code and benchmark suite}
}

@misc{deSousa2014,
title = {An Open Source IEC 61131-3 Integrated Development Environment},
ISSN = {1935-4576},
DOI = {10.1109/indin.2007.4384753},
booktitle = {2007 5th IEEE International Conference on Industrial Informatics},
publisher = {IEEE},
author = {Tisserant, Edouard and Bessard, Laurent and de Sousa, Mario},
year = {2007},
month = Jul,
pages = {183–187}
}

@inbook{Govil2017,
title = {On Ladder Logic Bombs in Industrial Control Systems},
ISBN = {9783319728179},
ISSN = {1611-3349},
DOI = {10.1007/978-3-319-72817-9_8},
booktitle = {Computer Security -- {ESORICS} 2017 International Workshops, {CyberICPS} 2017 and {SECPRE} 2017},
publisher = {Springer International Publishing},
author = {Govil, Naman and Agrawal, Anand and Tippenhauer, Nils Ole},
year = {2017},
month = sep,
address = {Oslo, Norway},
pages = {110--126},
}

@article{Rinieri2026,
title = {PLC-Defuser: Detecting hidden Ladder Logic Bombs in PLCs via Control Flow Graph and model checking},
volume = {169},
ISSN = {0167-4048},
DOI = {10.1016/j.cose.2026.104983},
journal = {Computers \& Security},
publisher = {Elsevier BV},
author = {Rinieri, Lorenzo and Iacobelli, Antonio and Melis, Andrea and Prandini, Marco and Callegati, Franco},
year = {2026},
month = Oct,
pages = {104983}
}

@misc{Rinieri2026artefact,
author = {Rinieri, Lorenzo and Iacobelli, Antonio and Melis, Andrea and Prandini, Marco and Callegati, Franco},
title = {{PLC\_Defuser}: tool and datasets (SWaT, GRFICS, Water\_tank)},
year = {2024},
version = {v1.0.0},
publisher = {Zenodo},
doi = {10.5281/zenodo.14014820},
note = {Archived v1.0.0 release; the SWaT corpus used for the linear-trigger tier of RQ6.}
}

@inproceedings{Zonouz2014arcade,
series = {ASE'12},
title = {Arcade.PLC: a verification platform for programmable logic controllers},
DOI = {10.1145/2351676.2351741},
booktitle = {Proceedings of the 27th IEEE/ACM International Conference on Automated Software Engineering},
publisher = {ACM},
author = {Biallas, Sebastian and Brauer, J\"{o}rg and Kowalewski, Stefan},
year = {2012},
address = {Essen, Germany},
month = Sep,
pages = {338–341},
collection = {ASE'12}
}

@inproceedings{Zhang2021attkfinder,
series = {RAID '21},
title = {{AttkFinder}: Discovering Attack Vectors in {PLC} Programs Using Information Flow Analysis},
DOI = {10.1145/3471621.3471864},
booktitle = {24th International Symposium on Research in Attacks, Intrusions and Defenses},
publisher = {ACM},
author = {Castellanos, John H. and Ochoa, Martin and Cardenas, Alvaro A. and Arden, Owen and Zhou, Jianying},
year = {2021},
month = oct,
address = {Donostia/San Sebasti\'{a}n, Spain},
pages = {235--250},
numpages = {16},
collection = {RAID '21},
}

\end{document}